\documentclass{new_tlp}
\usepackage{mathptmx}

\newcommand\bcmdtab{\noindent\bgroup\tabcolsep=0pt%
  \begin{tabular}{@{}p{10pc}@{}p{20pc}@{}}}
\newcommand\ecmdtab{\end{tabular}\egroup}

\newcommand\nCite[1]{\citeANP{#1} \shortcite{#1}}

\usepackage{amsmath}
\usepackage{graphicx}
\usepackage{multirow}
\usepackage{url}
\usepackage{algorithm}
\usepackage{algpseudocode}
\makeatletter
\renewcommand{\ALG@name}{Encoding}
\makeatother

\usepackage{booktabs}
\usepackage{graphicx}
\usepackage{subcaption}
\usepackage{bbding}

\algnewcommand{\LeftComment}[1]{\(\triangleright\) #1}
\usepackage{url}

\newcommand\INDSTATE{\State\hskip2em}

\jdate{Month year}
\pubyear{yyyy}
\pagerange{\pageref{firstpage}--\pageref{lastpage}}
\doi{S1471068401001193}

\title[Neuro-Symbolic AI for Compliance Checking]{Neuro-Symbolic AI for Compliance Checking\\ of Electrical Control Panels\thanks{Research partially supported by MISE (today MIMIT) under projects ``MAP4ID - Multipurpose Analytics Platform 4 Industrial Data” Proj. N. F/190138/01-03/X44 and MUR under PNRR  project PE0000013-FAIR, Spoke 9 - Green-aware AI – WP9.1. This work has been carried out while Alessandro Quarta was enrolled in the Italian National Doctorate on Artificial Intelligence run by Sapienza University of Rome with University of Calabria. A preliminary version appeared in CEUR Workshop Proceedings vol. 3204 pp.247--253.}}

\author[V. Barbara, M. Guarascio, N. Leone, G. Manco, A. Quarta, F. Ricca and E. Ritacco]
    {Vito Barbara\\
    University of Calabria, Italy\\
    \email{barbara.vito@unical.it}
    \and 
    Massimo Guarascio\\
    ICAR-CNR, Italy\\
    \email{massimo.guarascio@icar.cnr.it}
    \and
    Nicola Leone\\
    University of Calabria, Italy\\
    \email{nicola.leone@unical.it}
    \and
    Giuseppe Manco\\
    ICAR-CNR, Italy\\
    \email{giuseppe.manco@icar.cnr.it}
    \and 
    Alessandro Quarta\\
    Sapienza University of Rome, Italy\\
    \email{alessandro.quarta@uniroma1.it}
    \and
    Francesco Ricca\\
    University of Calabria, Italy\\
    \email{ricca@mat.unical.it}
    \and
    Ettore Ritacco\\
    University of Udine, Italy\\
    \email{ettore.ritacco@uniud.it}
    }

\begin{document}
\label{firstpage}
\maketitle

%\lefttitle{Barbara et al.}

\maketitle

\begin{abstract}
Artificial Intelligence plays a main role in supporting and improving smart manufacturing and Industry 4.0, by enabling the automation of different types of tasks manually performed by domain experts. In particular, assessing the compliance of a product with the relative schematic is a time-consuming and prone-to-error process.
In this paper, we address this problem in a specific industrial scenario. In particular, we define a Neuro-Symbolic approach for automating the compliance verification of the electrical control panels. Our approach is based on the combination of Deep Learning techniques with Answer Set Programming (ASP), and allows for identifying possible anomalies and errors in the final product even when a very limited amount of training data is available.
The experiments conducted on a real test case provided by an Italian Company operating in electrical control panel production demonstrate the effectiveness of the proposed approach.
\end{abstract}


\section{Introduction}
\label{sec:intro}\label{sec:contribution}

With the rise of new technologies, industry moved a step forward to a new era in the field of manufacturing. This complex transformation, including the integration of emerging paradigms and solutions such as \emph{Artificial Intelligence} (AI), \emph{Human-Computer Interaction}, \emph{Cloud Computing}, \emph{Industrial Internet Of Things} (IIoT) and \emph{Blockchain}, is referred as \emph{Industry 4.0}. The impact of the field is witnessed by the effort to promote its development within several national economic policies. For example, the Italian Ministry of Development (nowadays called Ministry of Enterprise and Made in Italy, and identified by the MIMIT acronym) is funding the application of AI to the manufacturing processes to improve efficiency and push the development and modernization of Italian SMEs.
In this evolving scenario, Quality Control (QC) is greatly benefiting from the adoption of advanced AI tools and techniques, that can allow for speeding up or automatizing processes of assessment about integrity, working capability, and durability of the products~\cite{AI-Industry40}. 
In particular, the automation of the compliance verification process for products is among the promising applications of AI for QC that poses a significant challenge for all manufacturing-related businesses, because it can make more efficient a necessary but costly and time-consuming operation in the supply chain.

Among the projects funded by MIMIT is the one titled ``Multipurpose Analytics Platform 4 Industrial Data'' (MAP4ID), where one of the main use cases is precisely the development of an AI capable of automating the compliance checking of Electrical Control Panels (ECPs).

Basically, an ECP is an enclosure, typically a metal or plastic box, which contains electrical components to control and monitor various mechanical processes, motors, sensors, and actuators. 
ECPs are employed to regulate a wide variety of components used in industry: \emph{e.g.}, they allow to control of mechanical equipment, electrical devices, manufacturing machinery, etc. 

One of the basic QC tasks in the manufacturing of ECPs requires checking the \textit{compliance} of the produced control panels with their schematics.
Automating this task is particularly relevant since it is currently manually performed by human experts, which makes the whole process inefficient, expensive, and prone to errors. The release of defective ECPs (due to poor quality control) can cause exposure to penalties by the customer and compromise the company's reputation.
The adoption of AI-based tools can greatly mitigate these risks by enabling continuous monitoring of the whole production chain and early detection of issues in each stage of the production process.

\paragraph{Main Problem.}
In this work, we devise an innovative approach combining Deep Learning (DL) \cite{DBLP:books/daglib/0040158}, and Answer Set Programming (ASP)~\cite{DBLP:journals/cacm/BrewkaET11,DBLP:journals/ngc/GelfondL91} to support the QC for the production of electrical control panels. Here, the main task consists in identifying anomalies in the final product, such as the lack, the misplacing, or the wrong connectivity of the electrical components in the cabinet of the ECP, by just analyzing an image of the assembled product.
Important requirements are that the AI must be capable of producing the results of the compliance-checking task in a very short time (in the order of seconds) and with high accuracy ($>$ 90\%), to enable the integration into a tool assisting human inspectors that delivers real-time and robust performance.

This problem is made very challenging for standard DL approaches by the following main issues:

\begin{enumerate}
    \item \textit{Data scarcity.} Although the companies can produce sufficient amounts of data, semantics, and labels are often missing from images. In particular, in our  scenario, such a problem affects both the data representations \emph{i.e.}, the pictures depicting the ECPs, and the correspondent schematics. Indeed, supervised information about the position, dimensions, and typology of the installed components is missing for the pictures. As regards the schematics, although they seem to provide a more detailed representation, the possibility of easily translating them into actionable constraints (in the form of grammar rules) strongly depends on the underlying software used to produce them.
    
    \item \textit{Custom Designs.} Despite ECP being made of standard components, there is no standard set of schematics for ECPs. Usually, the design of a solution is customized and very specific for the needs of a specific customer. Thus, the AI must be able to work with different schematics without requiring any additional training. 
\end{enumerate}

\paragraph{Contribution.}
In this work, we define a solution approach composed of two main phases:
\begin{enumerate}
    \item First, a Deep Learning based solution allows for recognizing the electrical components (object detection) from the images of the panels and reconstructing the scheme. In this phase, a number of data augmentation strategies are also exploited to cope with the lack of labeled data.
    \item Then, an Answer Set Programming-based system is used to compare the scheme reconstructed from the picture with its original schematic in order to discover possible mismatches/errors.
\end{enumerate}

The contribution of this paper can be located in the challenge of providing a suitable combination of learning and reasoning through the development of integrated components, which, nowadays, is identified by the buzzword \textit{neuro-symbolic AI}~\cite{DBLP:conf/aaaiss/GarcezBRFHIKLMS15}.
Actually, our system can be classified as a \textit{Neural|Symbolic} architecture (or architecture of Type 3) according to Henry Kautz's taxonomy~\cite{DBLP:journals/aim/Kautz22}, where DL is used for sensing (detecting components) and a reasoner (ASP-based) is used for checking conformance and detecting issues.

Although based on a conceptually straight combination of DL and ASP, an experiment conducted on (scarce) data provided by an Italian SME leader in the production of ECPs confirms that \textit{our neuro-symbolic system can deliver the expected performance}, which is the main acceptability criteria to be fulfilled by a successful real-world application. 

\section{Framework overview}
\label{sec:framework}

\begin{figure}[t!]
    \centering
	\includegraphics[width=.95\textwidth]{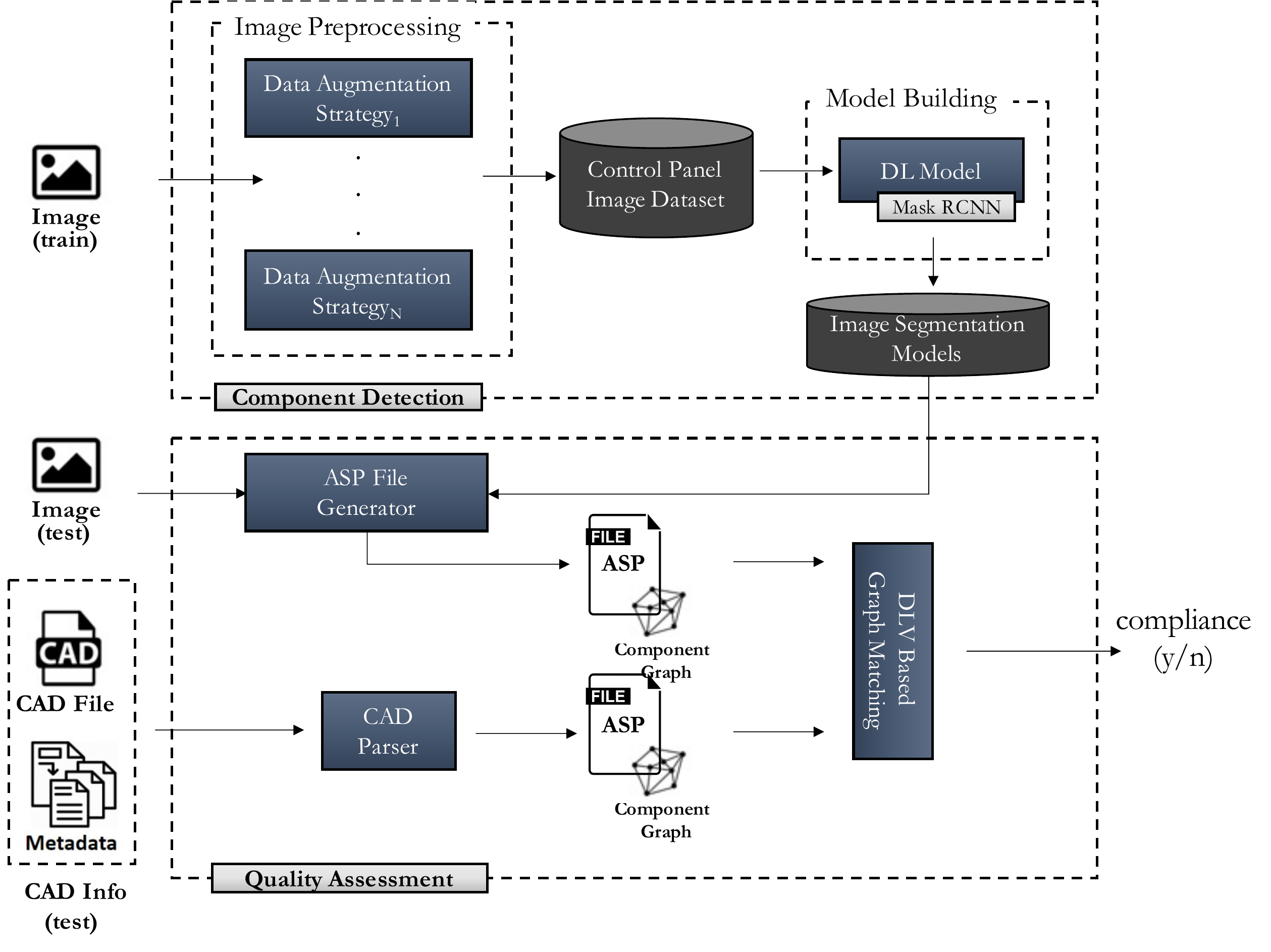}
	\caption{Framework for Automatic Compliance Verification.} 
	\label{fig:framework}
\end{figure}

In this section, we illustrate the solution approach devised to address the main problem of how to automate the compliance verification process of control panels. As highlighted in Section \ref{sec:intro}, an effective solution for this problem has to cope with the challenges of understanding image contents and extraction of the constraints encoded in schematics, while coping with the issues of lack of labeled data and unlabeled data distribution. 
To this aim, we defined the framework shown in Figure \ref{fig:framework} that includes two main macro-modules, respectively named \emph{Component Detection} and \emph{Quality Assessment}. 

The former is devoted to recognizing the electrical components assembled in the cabinet. Basically, it includes the modules characterizing the adopted machine learning methodology, whose main objective is to identify the components of the panel from a picture. Specifically, a set of \emph{Data Augmentation and generation} techniques builds a suitable dataset (robust to overfitting), which feeds a Convolutional Neural Network (CNN) based model trained to perform the component detection. 

The latter  exploits ASP to tackle the task of compliance checking. It automatically compares the control panel scheme built starting from the neural network output and the corresponding schematic to highlight any anomalies.

\section{Component Detection via Deep Learning}
\label{sec:component_detection}

The component detection is meant to recognize, given a picture representing a panel, the type and geometric position of each component within the panel. This is a preliminary and fundamental step since, in order to check the  compliance of the cabinets with their schematics, we need first to understand their composition. The main problem in this step is given by the scarcity of data, as well as the lack of labeling annotations. This is a typical scenario characterizing industrial processes: the  quality of a machine learning model relies on the data used to train it; however, the latter requires an accurate labeling process that is time and resource-consuming and hence difficult to obtain. 
In our framework, we address these issues by exploiting a synthetic data generation process that allows us to enrich the starting training set. 

\begin{figure*}[t!]
    \centering
    \subfloat[Control panel background.\label{fig:background}]{%
    \includegraphics[trim = 1 20 2 48, clip, width=0.35\textwidth]{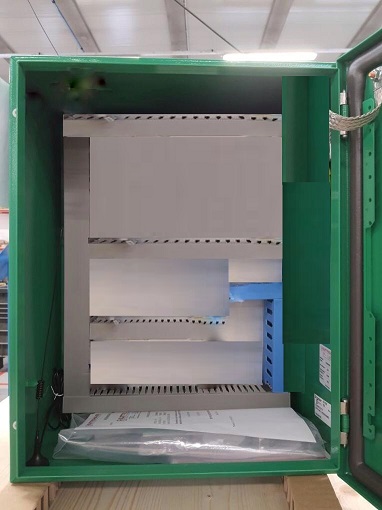}%
    }\label{fig:empty_panel}
    $\quad\quad$
    \subfloat[Generated image with a mix of large and small components.
    \label{fig:synth1}]{%
    \includegraphics[trim = 1 30 2 100, clip, width=0.35\textwidth]{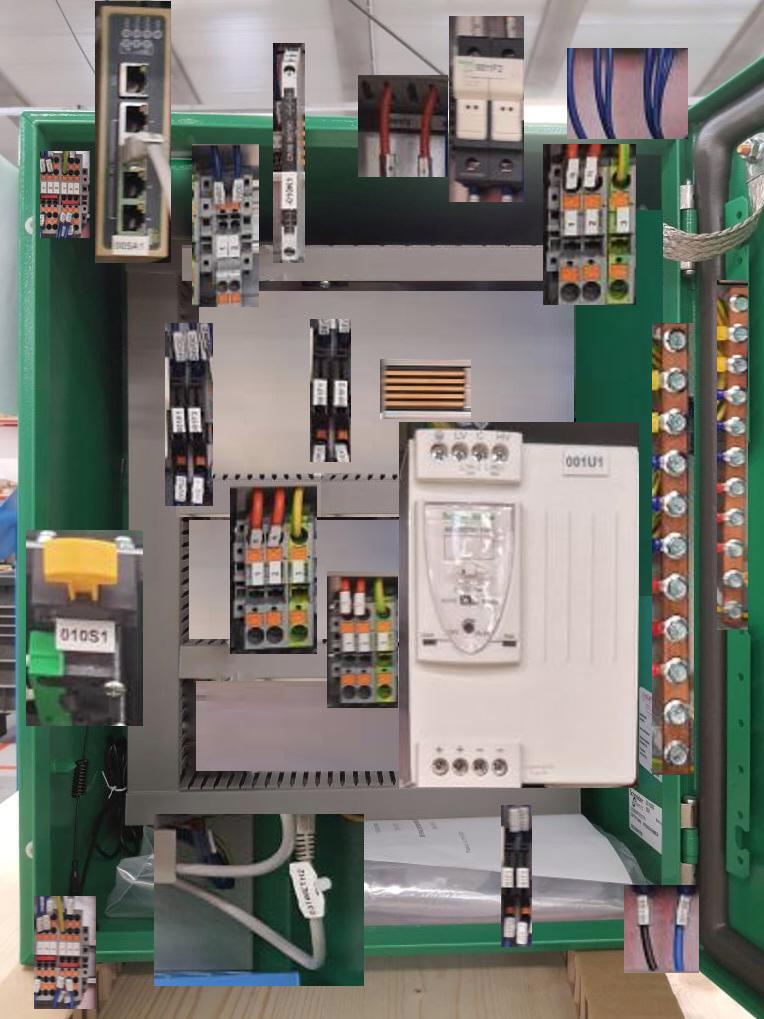}%
    }
    %\\
    %\vspace*{1em}
    $\quad\quad$
    \subfloat[
    Generated image with small components and noisy elements.
    \label{fig:synth2}]{%
    \includegraphics[trim = 1 30 2 100, clip, width=0.35\textwidth]{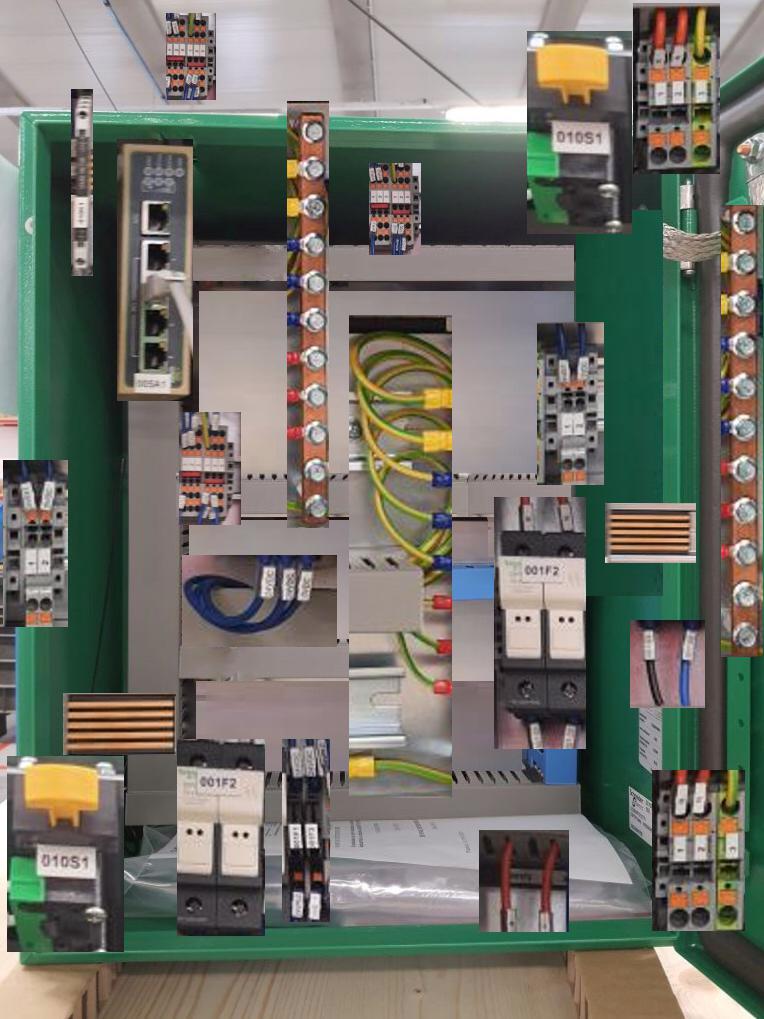}%
    }
    $\quad\quad$
    \subfloat[Mask-RCNN output (on a real image).\label{fig:output}]{%
    \includegraphics[width=0.35\textwidth]{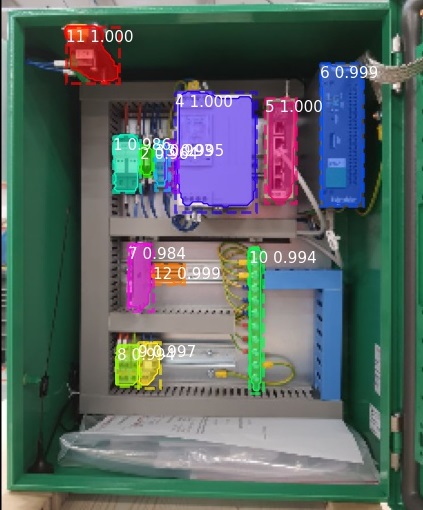}%
    }
\caption{Input and output of the component detection approach.}
\label{fig:lb_exp}
\end{figure*}

\subsection{Data Augmentation and Generation}
\label{sec:preprocessing}

Basically, our synthetic data generation method is fed with three inputs: (\textit{i}) a picture showing an empty cabinet, (\textit{ii}) a catalog including all the available components that can be installed in a cabinet, and (\textit{iii}) a limited number of real pictures that will be manipulated in order to add noisy elements in the generated data, by exploiting a suitable strategy described in the following.

The core idea is to enrich the ground truth (consisting of a limited number of images) with synthetic images, where the area of the empty cabinet is filled with random components picked from the catalog. Notice that, at this stage, we are not interested in generating compliant panels, since our only objective at this stage is to build a suitable object detector that is capable of recognizing both the component and its geometrical position and extension. The size of the catalog and the randomness of the composition allow us to generate a suitable number of images where each component can be included with a suitable frequency, thus making the result dataset robust to object detection and segmentation learning tasks. 

This simple approach can be further combined with other image augmentation strategies (\emph{Image Processing} module in Figure~\ref{fig:framework}) with the aim of yielding a training set that includes a sufficient and diversified number of examples for learning the model. 
In particular, our framework also includes traditional data augmentation strategies \emph{i.e.}, \emph{Gaussian Blur} and \emph{PerspectiveTransform}. As regards the former, the idea consists in introducing imperfections into data so as to make component detection more resilient to data changes, it is obtained by averaging contiguous pixel values. The last one allows for applying random four-point perspective transformations to images.

The resulting dataset from this process will include all the necessary features for the training phase: (\textit{i}) a large number of different pictures, (\textit{ii}) the position of each component, (\textit{iii}) the type of each component. Notably, since each component depicted in the synthetic pictures  is randomly placed, the detection model will be forced to learn the intrinsic features of each component, instead of considering positional features, that may vary in the different schematics.
Within a cabinet, there are other ``auxiliary'' elements that are simplified in a schematic, mainly separation boxes, metal runners, and cables. For simplicity, we call them noise to randomly add to the generated images in order to make the object detection model able to distinguish and ignore these elements.

Figure~\ref{fig:lb_exp}, show some examples of the data generation process. We can observe the empty cabinet (Figure~\ref{fig:background}), and two instances where it is filled with random components (Figures~\ref{fig:synth1} and~\ref{fig:synth2}). Notice that the synthetic data does not necessarily represent a realistic situation. As already mentioned, this is not an issue since our purpose here is to strengthen the object detection and segmentation phase, which is discussed below.
% subsection. 

\subsection{Component Detection
}
\label{sec:model}

\begin{figure*}[t!]
    \centering
	\includegraphics[width=1.\textwidth]{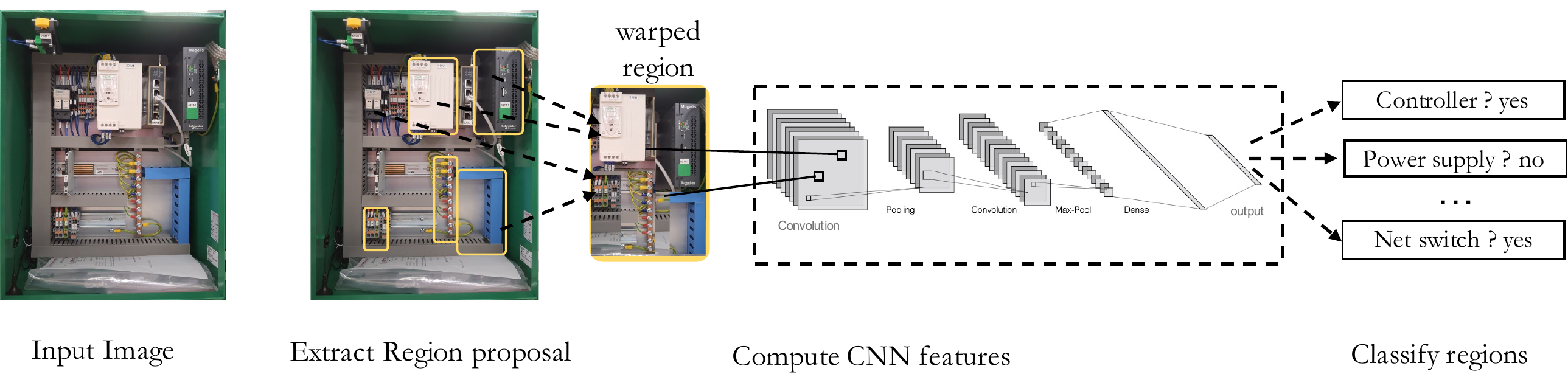}
	\caption{R-CNN working flow.} 
	\label{fig:rcnn_concept}
\end{figure*}

The \emph{Model Building} module in Figure~\ref{fig:framework} allows for training the deep architecture used to perform the component detection. For this, we adopted the \emph{Mask R-CNN} convolutional neural architecture proposed by \nCite{He17}.
In general, R-CNN (Region based CNN) refers to a family of neural architectures adopting a \emph{Multi-shot} approach. The underlying idea is a two-step process: first, different bounding boxes across possible regions of interest (RoIs) are extracted; then, such regions are independently evaluated through a CNN architecture in order to map them to any of the proposed classes (see Figure~\ref{fig:rcnn_concept}). 

Mask R-CNN extends a specific architecture named \emph{Faster R-CNN} \cite{faster_rcnn} that includes two main components: \textit{(i)} \emph{Region Proposal Network} (RPN), a deep neural network aimed at extracting RoIs from the picture, and \textit{(ii)} \emph{Fast R-CNN}, a neural architecture that performs classification, by scaling a region to a predefined size thus enabling the computation of a set of CNN feature maps. 
The main advantage of the Faster R-CNN architecture is a suitable trade-off between competitive accuracy in terms of object recognition, and relative speed in the recognition phase. By contrast, other approaches based on Single-Shot architectures such as YOLO \cite{Redmon_2016_CVPR} or SSD \cite{Liu16} focus on fast recognition, at the cost of recognition accuracy. This is clearly not acceptable in our scenario, where we aim at checking compliance, and missing a component in the picture would result in a failure in the check. 
Mask R-CNN further improves Faster R-CNN by introducing a further branch for predicting segmentation masks on each Region of Interest. The recognition of the mask is crucial in our scenario since it allows to precisely identify the geometrical position of the component within the panel. Technically, Mask R-CNN rebuilds the mask by resorting to an alignment component and a mask head, composed of two convolutional layers and capable of generating a mask for each RoI in order to segment the picture in a pixel-to-pixel fashion.

Mask R-CNN relies on a backbone convolutional architecture. In our framework, we used ResNet (Residual Network) \cite{residualNets}, a very deep CNN architecture characterized by \emph{residual blocks} and \emph{skip connections}.
These two features guarantee both, fast convergence in the training stage and expressiveness/accuracy in the recognition phase. 
We further strengthened the training phase by exploiting  \emph{Transfer Learning}. 
In particular, we used a ResNet pre-trained on \emph{COCO} dataset%
\footnote{Available online at: \url{https://cocodataset.org/\#home} [Last Accessed: June 2022].}, which was also fine-tuned for our specific scenario, by exploiting the generated dataset with the artificially generated labeled components.

Figure~\ref{fig:output} shows the output of the recognition phase, on a real picture representing a true panel. We can see that the model successfully recognizes all available components, and additionally devises a contour of their geometric extension. These contours represent one of the inputs to the reasoning component.

\section{Compliance-checking in Answer Set Programming}
\label{sec:asp_component}
In this section, we describe the reasoning component of our architecture for compliance checking.
In particular, this component has been implemented by resorting to Answer Set Programming (ASP).
ASP is a well-established paradigm for declarative programming and non-monotonic reasoning developed in the area of Knowledge Representation and Reasoning~\cite{Baral:2003:KRR:582493,DBLP:journals/cacm/BrewkaET11,DBLP:conf/agp/BonattiCLR10,DBLP:journals/ngc/GelfondL91}.
ASP has been employed to develop many academic and industrial applications of AI ~\cite{DBLP:journals/aim/ErdemGL16,DBLP:journals/tplp/GebserMR20,DBLP:journals/ai/CalimeriGMR16,DBLP:conf/lpnmr/GrassoILR09,DBLP:conf/birthday/GrassoLMR11,DBLP:journals/tplp/DodaroGLMRS16}.
ASP is based on logic programming and non-monotonic reasoning, and it allows for flexible declarative modeling of search problems, by means of logic programs (collection of rules), whose intended models (answer sets) encode solutions~\cite{Baral:2003:KRR:582493,DBLP:journals/cacm/BrewkaET11}. 
The specification (logic program) described in the following can be fed to an ASP system to actually compute the solutions to the modeled program~\cite{DBLP:journals/aim/LierlerMR16}. 

The reasoning module is fed by two handlers, named \emph{ASP File Generator} and \emph{CAD Parser}. The former component is devoted to translating the objects recognized by the neural model in ASP facts (a file containing a list of facts concerning coordinates and membership of the electrical component), similarly, the second one yields a list of facts from the input CAD image. 

In the following, we focus on the core parts of our solution and simplify some technical aspects that do not impact the comprehension of the working principle of our solution. This is done with  the aim of making the presentation more accessible and meeting space requirements.
Hereafter, we assume the reader to be familiar with ASP. For  details please refer to~\cite{DBLP:journals/cacm/BrewkaET11,Baral:2003:KRR:582493,DBLP:journals/ngc/GelfondL91}.

\subsection{Input specification}
In ASP the input specification is made by a set of ``facts'', which are assertions that model true sentences. Thus, the labeled schematic of the circuit (we informally refer to it as cad), and the output of the Mask R-CNN net (exemplified in Figure \ref{fig:output}) are converted in a set of ASP facts of the following form:

\begin{verbatim}
object(LABEL, ID, X_TOP_L, Y_TOP_L, X_BOT_R, Y_BOT_R, MEMBERSHIP).
\end{verbatim}

These facts provide information about the components like their label, id, top-left and bottom-right coordinates, and membership.
In particular, the membership is valued with ``\texttt{cad}'' if the object modeled is part of the schematic of the panel, and ``\texttt{net}'' if it is recognized by the neural network in the actual picture we are comparing to the schematic.
Moreover, we also compute a graph of topological relations among objects, providing information on relative position and distance among objects. The relative position and the distance among components are actually calculated by our ASP program, but for simplicity, we assume here they are given in input as facts of the form:

\begin{verbatim}
between(ID, START_ID, END_ID, DIR, MEM).
manhattan(ID1, ID2, DIST, MEM1, MEM2).
\end{verbatim}

The \verb$between$ predicate denotes the neighbors for the component \verb$ID$ along the direction \verb$DIR$ having \verb$MEM$ as membership; additionally, the \verb$manhattan$ predicate  specifies the manhattan distance between the two components \verb$ID1$ and \verb$ID2$, where the terms \verb$MEM1$ and \verb$MEM2$ stand for their membership.

\begin{algorithm}[t!]
%\footnotesize
\caption{ASP program modeling compliance}\label{alg:ASP}
\begin{algorithmic}[1]
\State  \LeftComment{Calculate auxiliary information}
\INDSTATE \verb|previous(ID, Start_ID, D, M):- between(ID, Start_ID, _ , D, M).|
\INDSTATE \verb|after(ID, End_ID, D, M):- between(ID, _, End_ID, D, M).|

\State \LeftComment{Guess mapping between cad components and net components}
\INDSTATE \verb|simpObject(C1,ID1,M) :- object(C1,ID1,_,_,_,_,M).|
\INDSTATE \verb$mapped(ID1,ID2) || noMapped(ID1,ID2)$
\INDSTATE \hspace{1cm} \verb$:- simpObject(C1,ID1,"cad"),simpObject(C1,ID2,"net").$

\State \LeftComment{ No element from the cad is mapped twice}
\INDSTATE \verb|:- mapped(Cad_ID,Net_ID1), mapped(Cad_ID,Net_ID2),| 
\INDSTATE \ \ \ \ \verb|Net_ID1!=Net_ID2.|

\State \LeftComment{ No element from the net is mapped twice}
\INDSTATE \verb|:- mapped(Cad_ID1,Net_ID), mapped(Cad_ID2,Net_ID),| 
\INDSTATE \ \ \ \ \verb|Cad_ID1!=Cad_ID2.|

\State \LeftComment{ Minimize the cad elements without a mapping}
\INDSTATE \verb|atLeastOne(Cad_ID) :- mapped(Cad_ID,_).| 
\INDSTATE  \verb|:|\url{~} \verb|simpObject(C1,ID1,"cad"), not atLeastOne(ID1). [1@3,C1,ID1]|

\State \LeftComment{ Optimize mapping by relative position}
\INDSTATE \verb|:|\url{~} \verb|mapped(Cad_ID1, Net_ID1), mapped(Cad_ID2,Net_ID2),|
\INDSTATE \ \ \ \ \verb|previous(Cad_ID1,Cad_ID2,DIR,"cad"),| 
\INDSTATE \ \ \ \ \verb|not previous(Net_ID1, Net_ID2, DIR,"net").| 
\INDSTATE \ \ \ \ \verb|[1@2,Cad_ID1, Net_ID1,Cad_ID2,Net_ID2,DIR]|
\INDSTATE \verb|:|\url{~} \verb|mapped(Cad_ID1,Net_ID1), mapped(Cad_ID2,Net_ID2),| 
\INDSTATE \ \ \ \ \verb|after(Cad_ID1, Cad_ID2,DIR,"cad"),|
\INDSTATE \ \ \ \ \verb|not after(Net_ID1,Net_ID2,DIR,"net").| 
\INDSTATE \ \ \ \ \verb|[1@2,Cad_ID1,Net_ID1,Cad_ID2,Net_ID2,DIR]|
\INDSTATE \verb|:|\url{~} \verb|mapped(Cad_ID1, Net_ID1),|
\INDSTATE \ \ \ \ \verb|previous(Cad_ID1, Cad_ID2, DIR,"cad"),| 
\INDSTATE \ \ \ \ \verb|absent(_,Cad_ID2). [1@2,Cad_ID1,Net_ID1,Cad_ID2,DIR]|
\INDSTATE \verb|:|\url{~} \verb|mapped(Cad_ID1, Net_ID1),| 
\INDSTATE \ \ \ \ \verb|after(Cad_ID1, Cad_ID2, DIR,"cad"),| 
\INDSTATE \ \ \ \ \verb|absent(_,Cad_ID2). [1@2,Cad_ID1,Net_ID1,Cad_ID2,DIR]|

\State \LeftComment{  Optimize mapping by distance}
\INDSTATE \verb|:|\url{~} \verb|mapped(Cad_ID, Net_ID),|
\INDSTATE \ \ \ \ \verb|manhattan(Cad_ID, Net_ID, Dis,"cad","net").|  
\INDSTATE \ \ \ \ \verb|[Dis@1,Cad_ID,Net_ID,Dis]|
\State \LeftComment{ Identify absent and in excess components}
\INDSTATE \verb|mappedCad(ID1):- mapped(ID1,_).|
\INDSTATE \verb|mappedNet(ID1):- mapped(_,ID1).|
\INDSTATE \verb|absent(C1,ID1):- simpObject(C1,ID1,"cad"), not mappedCad(ID1).|
\INDSTATE \verb|excess(C1,ID1):- simpObject(C1,ID1,"net"), not mappedNet(ID1).|
\end{algorithmic}
\end{algorithm}

\subsection{ASP program}
We now present ASP program (see Encoding ~\ref{alg:ASP}) that encodes in a uniform way (w.r.t. the input instance provided as a set of facts) the compliance problem.
First, the graph is preprocessed (lines 2-3), by calculating useful information about the relative positions of the objects.
Next, according to the “guess-and-check” programming methodology, a disjunctive rule guesses the mapping between ``\texttt{cad}'' components of the schematic and ``\texttt{net}'' components predicted by the neural network (see lines 6-7).

The disjunctive rule can be read as follows: `Given a cad component and a net component of the same type, the two can be mapped, or not''.
The candidate solutions are filtered out by the constraints in lines 9-13, ensuring that the same element of the cad is not mapped twice, and the same element of the net is not mapped twice.

The optimal mapping is obtained by weak constraints in lines 15-35. 
In detail, the program first minimizes the cad elements without a mapping (lines 15-16), then (also in order of priority) the weak constraints in lines 18-31 ensure that ``If a \texttt{cad} component \verb$ID1$ is mapped to a \texttt{net} component \verb$ID2$, \verb$ID1$ neighbors should be mapped to \verb$ID2$ neighbors''.

The mapping is further optimized considering the distance (lines 33-35) between \texttt{cad} components and \texttt{net} components. 
The distance is optimal when the elements are in the same position in \verb$"net"$ and \verb$"cad"$.
Finally, the program identifies components that are absent or in excess w.r.t. the schematic by rules in lines 37-40.

\section{Evaluation}
\label{sec:experiments}

This section describes a suite of experiments  we conducted,  devoted to demonstrating the effectiveness of our approach and its suitability for the industrial scenario. Specifically, we are interested in evaluating the capability of DL-based approach in recognizing the cabinet components when no training data are available, and the scalability of the ASP-based technique in verifying the conformance of the image with the schematics. 

\begin{figure*}[t]
\centering
\subfloat[Averaged PR-Curve.\label{fig:pr-curve}]{%
  \includegraphics[width=0.45\textwidth]{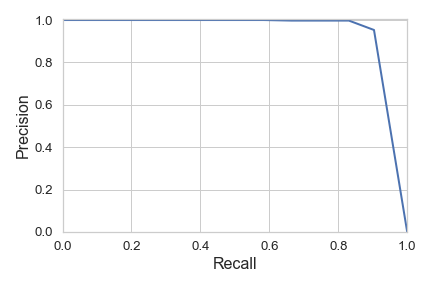}%
}
\hspace{1.cm}
\subfloat[PR-Curves computed per instance.\label{fig:pr-curve-per-instance}]{%
  \includegraphics[width=0.45\textwidth]{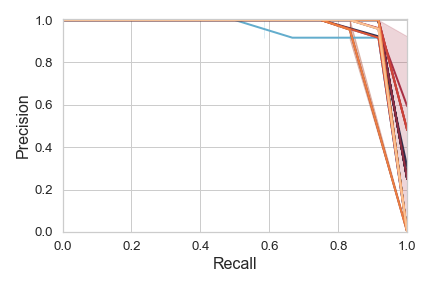}%
}
\caption{Precision-Recall Curves on test set.}
\label{fig:prcurves}
\end{figure*}

\subsection{Experimental setup}

We set up the experimentation by considering the extreme scenario where no labeled examples are available. Therefore, our training set used includes only the synthetically generated images (by using the data augmentation techniques described in section~\ref{sec:component_detection}), while the real pictures of the EPCs are used to evaluate the predictive performances. The final result of this process is a training set composed of $\sim10,000$ colored images synthetically generated with size ($320 \times 320$) and a test set of $32$ images depicting real control panels with the same size as the training ones.

The model discussed in Section \ref{sec:component_detection} has been implemented in the form of a python prototype based on TensorFlow\footnote{TensorFlow machine learning library: \url{https://www.tensorflow.org/} [Last Accessed: June 2022].} library. The experiments were performed on an NVidia DGX Station equipped with 4 GPU V100 32GB. As described in section \ref{sec:component_detection}, a ResNet instance (including 101 layers) is used as the backbone of the component detection model, the Mask R-CNN, that is trained over $200$ epochs with $batch\_size = 2$, while \emph{Adam} is adopted as optimizer with learning rate $\mathit{lr}= 10^{-4}$. 

To assess the capability of the proposed approach in detecting the components installed in the ECPs, a number of traditional measures and well-known metrics for the Object Recognition tasks have been used. In this sub-section, we briefly introduce and define such measures. 
The first measures we consider are the standard Precision and Recall metrics, defined as $p = \frac{TP}{TP + FP}$ and $r = \frac{TP}{TP + FN}$. Here, $TP$, $FP$,  $FN$, and $TN$ denote respectively the number of cases that are: positive and correctly classified, positive and incorrectly classified, negative and incorrectly classified, and negative and correctly classified. Hence, a Precision-Recall Curve can be obtained by computing and plotting the precision against the recall for different threshold values (\emph{i.e.,} the detection probabilities of the model). 

In an object detection scenario, precision and recall represent  the capability of the prediction model to identify the boxes that contain the target objects. In particular, for a given object the focus is on comparing the true bounding box with the predicted bounding box, and the $TP$, $FP$,  $FN$, and $TN$ values depend on the degree of overlap between these two boxes. Given two boxes, the \emph{Intersection Over Union} ($IoU$) is defined as the fraction of the overlapping area between the ground truth $b$ and the predicted bounding box $\hat{b}$:
\begin{equation}
    \mathit{IoU}(b,\hat{b}) = \frac{b \cap \hat{b}}{b \cup \hat{b}}
\end{equation}
Then, given a threshold $\theta$, an object with a true bounding box $b$ and a predicted bounding box $\hat{b}$ is \emph{positive} if $\mathit{IoU}(b,\hat{b})>\theta$, and \emph{negative} otherwise.
For a given $\theta$, it is possible to devise a precision-recall curve by plotting all $p$/$r$ values relative to all objects and interpolating the resulting curve~\cite{he2017,faster_rcnn}.

Since the values of precision and recall are defined on a given $\theta$ threshold, we can define~\cite{he2017,faster_rcnn} \emph{Average Precision} and \emph{Recall} as the area represented by integrating over all possible thresholds: 
\begin{equation}
    \mathit{AP} = \int_{0}^{1} p(\theta) \, d\theta
    \quad \mbox{(and resp.)} \quad
    \mathit{AR} = \int_{0}^{1} r(\theta) \, d\theta \, .
\end{equation}
Finally, by averaging $\mathit{AP}$ (resp. $\mathit{AR}$) over all class components  we can finally obtain the \emph{mean average precision} ($\mathit{mAP}$) and \emph{mean average recall} ($\mathit{mAR}$) measures.

\begin{figure}[t!]
    \centering %[trim={left bottom right top},clip]
	\includegraphics[width=.35\textwidth,trim ={130 10 110 10},clip]{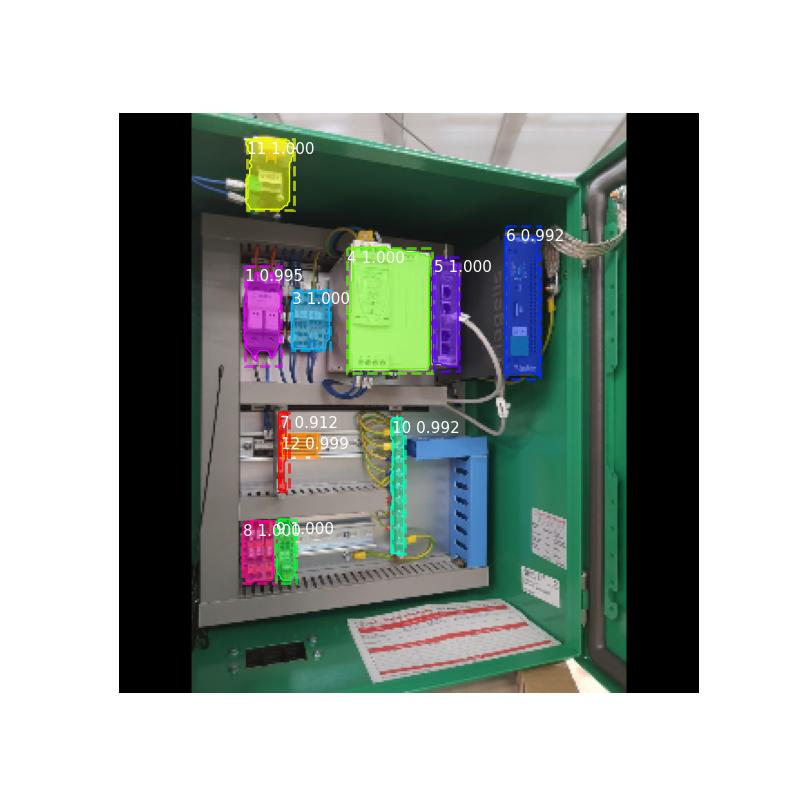}
	\caption{An example of inaccurate image acquisition. The side perspective of the image does not match the component images in the catalog leading to inaccurate predictions.} 
	\label{fig:wrong_prediction}
\end{figure}

\subsection{Evaluation results}

Here, we discuss the results in terms of the effectiveness of the DL-Based detection model and the scalability of the ASP module. 
For the first aspect,  the detection model exhibits optimal performances for both the quality measures, exhibiting values of $\mathit{mAP}=0.954$ and $\mathit{mAR}=0.935$. In order to evaluate the operational applicability of our approach in a real scenario, we conducted a further analysis by considering the values of precision and recall on a fixed $\mathit{IoU}$ threshold $\theta=0.5$. 
Basically, in this test case, precision and recall, respectively, represent the capability of the model to correctly recognize the components depicted in the picture and the percentage of recognized components w.r.t. the ground truth. 

Figure \ref{fig:pr-curve} reports the resulting precision-recall curve. The resulting area is $0.947$, which denotes a good performance of the detection model also considering the operational case. Figure \ref{fig:pr-curve-per-instance} shows a more detailed picture of the model performances by plotting the pr-curve for each instance. As expected, for almost all instances, the yielded curves highlight the good predictive accuracy of the model in recognizing the different types of components, except for one case in which the quality is slightly lower.  
The above evaluation shows that the component detection module is effective in recognizing the components of a panel: in particular, prediction errors can occur in rare cases with inaccurate image acquisition (\emph{e.g.}, non-frontal framing or inclusion of elements external to the cabinet) as the catalog provides only a limited number of component perspectives. An example of such behavior is depicted in Figure \ref{fig:wrong_prediction}, where accuracy is affected by the wrong perspective of the image. Since the ASP program performs the compliance task with optimal accuracy in our benchmark images, the accuracy of the system corresponds with the one of the neural model.

\begin{figure*}[t!]
    \centering
	\includegraphics[width=\textwidth
	]{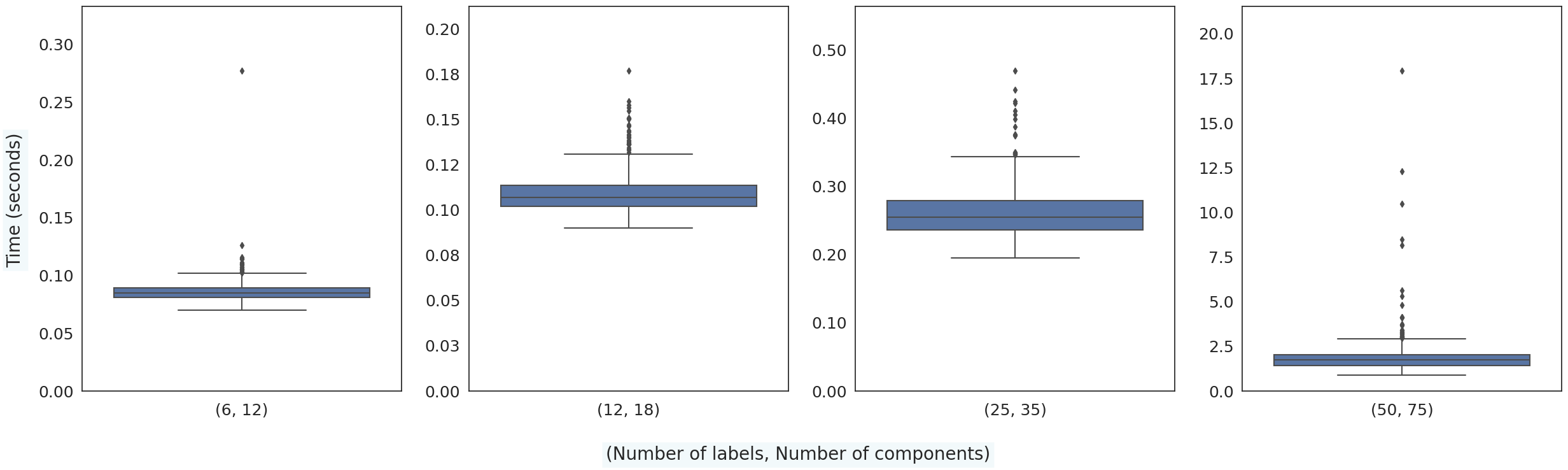}
	\caption{Performance of the ASP-Based component (Execution time).} 
	\label{fig:results-asp}
\end{figure*}

%%% ASP SECTION  
One might wonder whether the ASP component is efficient thus in a further experiment the execution time of the ASP-based component was measured.
We generated instances of compliance testing in a range of 6 to 50 labels (types of components), and of 12 to 75 components and averaged over 500 samples the execution time needed by our ASP engine DLV2 \cite{DBLP:conf/lpnmr/AlvianoCDFLPRVZ17} to solve the instances. The results reported in Figure~\ref{fig:results-asp} show that our system provides answers in a short time, in the order of seconds for instances sized as real-world ones, and performance is acceptable (avg. 1.93s, max about 18s) also for instances of 75 components. 
 
\section{Related works}
\label{sec:related}
In this section, we survey some relevant works that try to address the product quality assurance problem by leveraging AI-based strategies, then we review some preliminary works proposing solutions to integrate ML techniques with logic programming. 

\begin{table*}[b!]
\caption{Comparison of the ML/DL based approaches.}
\centering
%\footnotesize

\resizebox{.99\textwidth}{!}{
\begin{tabular}{p{15mm} p{25mm} p{40mm} p{20mm} p{8mm} p{10mm}}
\toprule
\textbf{Article}  &  
\textbf{Application \quad Scenario}  &  
\textbf{Solution} & 
\textbf{ML/DL Model(s)} &
\textbf{Neur. Symb.} &
\textbf{Labeled Data} \\

\midrule
\cite{tanuska2021smart} &
IIoT, Sensors data &
Combining Neural Networks and Sound Analysis &
Feedforward Neural Networks &
$\times$ &
\Checkmark \\

%\midrule
\cite{SCHMITT2020101101} &
Surface Mount technology manufacturing &
Merging ML with Edge Cloud Computing &
Traditional ML Supervised techniques &
$\times$ &
\Checkmark \\

%\midrule
\cite{Banus2021} &
Quality Control (Food Packages) &
Computer Vision &
ResNet, VGGNet, DenseNet &
$\times$ &
\Checkmark \\

%\midrule
\cite{Villa19} &
Quality Control (Printing Industry) &
Computer Vision &
CNN (loosely inspired to AlexNet) &
$\times$ &
\Checkmark \\

%\midrule
\cite{subakti18} &
Smart Factory (Machine Recognition) &
Combination of Computer Vision and Augmented Reality &
MobileNet &
$\times$ &
\Checkmark \\

%\midrule
\emph{Our Solution} &
ECP Compliance Verification &
A framework based on DL and Answer Set Programming &
Mask R-CNN &
\Checkmark &
$\times$ \\

\bottomrule
\end{tabular}
}
\label{table:rw_comparison}
\end{table*}

\paragraph{Compliance Checking through Machine Learning.} To the best of our knowledge, the problem of assessing the compliance of a product with its schematic through Artificial Intelligence techniques is new and quite unexplored however, some recent works tried to tackle similar tasks, in particular within \emph{Predictive Maintenance} field. For instance, \nCite{tanuska2021smart} propose a comprehensive framework integrating Industrial Internet of Things (IIoT) devices, neural networks, and sound analysis for detecting anomalies in the production chain.
\nCite{SCHMITT2020101101} define a holistic solution for quality inspection based on merging Machine Learning techniques and Edge Cloud Computing technology. A Deep Learning based approach for monitoring the process of sealing and closure of matrix-shaped thermoforming food packages is proposed by \nCite{Banus2021}. Specifically, Computer Vision techniques are exploited to process the images and perform quality checking. A comparison analysis performed by ranging different Convolutional Neural Network architectures (\emph{e.g.}, ResNet50, VGG19, ImageNet, etc.) highlights the best solutions to address this task. \nCite{Villa19} propose a  deep neural network (DNN) soft sensor enabling fast quality control for the Printing Industry. Basically, the solution allows for comparing the scanned surface of the print with the correspondent file that generated it and performs an automatic quality control process by learning features through exposure to training data. 
\nCite{subakti18} define and develop a deep learning-based framework to detect/recognize different machines and portions of machines for smart factories. MobileNets is used as the backbone for the machine recognition model, and it is deployed on mobile devices to support the operators in performing the machine classification through an augmented reality system. Experimental results on a real scenario show the capability of the approach in recognizing different machines and providing intuitive visualizations.

In Table \ref{table:rw_comparison}, we compare the main approaches proposed in the literature and highlight the differences w.r.t. our solution. The main advantage of our approach (the only one based on a neuro-symbolic architecture) stays in the nature of the symbolic component that does not require additional training to deal with new (unseen) schematics. Another distinguishing feature is the ability to work with data scarcity (i.e., small training sets).

\paragraph{ML and ASP integration.}
The integration of inductive with deductive reasoning is an emerging problem in Artificial Intelligence (AI). Several proposals were made to implement the reasoning process in complex deep neural network (DNN) architectures \cite{kathryn2018tensorlog,rocktaschel2017deep,yang2020neurasp,lin-etal-2019-kagnet,10.5555/3172077.3172109}.
The integration of deductive logical reasoning with the Deep Learning paradigm is a novel and quite unexplored research topic, although some recent works introduced interesting preliminary solutions~\cite{EbrahimiEBH21}. 
Concerning ASP, we recall that it is a declarative rule-based programming paradigm for knowledge representation and declarative problem-solving, that is known to be appropriate for executing complex knowledge-based applications \cite{DBLP:journals/aim/ErdemGL16}.
One of the main issues is to incorporate high-dimensional vector space and pre-trained models for perception tasks as handled in deep learning, which limits the applicability of ASP in many practical applications involving data and uncertainty.
Nonetheless, to overcome this issue a blending ASP with DL has been recently studied~\cite{yang2020neurasp}.

%\vspace{-0.2cm}
\section{Conclusions and future works}
\label{sec:conclusions}

Quality Control is a manually performed and prone-to-error task crucial for each company, indeed the release of defective products can damage the company's reputation and lead to the payment of penalties to the customer. 

This paper describes a Neuro-symbolic approach to checking the compliance of electrical control panels with their schematics. A picture of a control panel is fed as input to a neural network to recognize the installed components and their locations, then an ASP-based module is used to compare the scheme reconstructed from the picture with its original blueprint and detect possible mismatches/errors. 

The system can handle the lack of labeled data and is resilient to noise and variety in the specifications of schematics (no additional training is required, just an updated logical representation of the schematic). 
The overall system has been exploited in a practical use case provided by an italian SME leader in the production of ECPs, where it has been shown to fulfill the requirements both in terms of accuracy and evaluation time. 

Despite its practical utility, there is still room for improving the proposed framework. In fact, we plan to extend it along two research directions. Concerning the model, we can improve the learning phase by adopting a \emph{Triplet Loss} \cite{triplet} architecture and by changing the  model backbone (\emph{e.g.}, by resorting to \emph{Vision Transformers} \cite{ViT}). 
Another potential issue is that the proposed model disregards the depth of the cabinet. In practice, we only consider a two-dimensional model where each component is placed on a plane. There are situations, however, where components partially overlap frontally but occupy different positions in depth. For these situations, a more accurate model that also addresses depth estimation should be considered. 

The second line for possible  is represented by the reasoning modules, where the logic programs can be calibrated to compute suggestions for the user, as well as suggest alternate schematic plans. Finally, one could study whether a tighter integration of the neural and logic-based components can enhance the results provided by the vision procedure. 

\paragraph{Acknowledgments.} The authors express sincere gratitude to the anonymous referees for their valuable suggestions, which contributed to the improvement of our work. We would also like to acknowledge the exceptional support of Dimitri Buelli, Stefano Ierace, Salvatore Iiritano, Giovanni Laboccetta, and Valerio Pesenti elapsed during the development of the system presented in this paper. Their expertise and dedication have been instrumental in shaping the success of our research.

%\vspace{-0.15cm}
%\bibliographystyle{tlplike}
\bibliographystyle{acmtrans}
%\bibliography{references}

\end{document}